\documentclass[conference]{IEEEtran}
\IEEEoverridecommandlockouts
\usepackage{cite}
\usepackage{amsmath,amssymb,amsfonts,amsthm}
\usepackage{algorithmic}
\usepackage{graphicx}
\usepackage{textcomp}
\usepackage{xcolor}
\usepackage{booktabs}
\usepackage{url}
\usepackage{multirow}
\def\BibTeX{{\rm B\kern-.05em{\sc i\kern-.025em b}\kern-.08em
    T\kern-.1667em\lower.7ex\hbox{E}\kern-.125emX}}
\newcommand{\same}{$0^{\dagger}$}
\newtheorem{lemma}{Lemma}
\newtheorem{theorem}{Theorem}
\makeatletter
\newcommand{\linebreakand}{%
  \end{@IEEEauthorhalign}
  \hfill\mbox{}\par
  \mbox{}\hfill\begin{@IEEEauthorhalign}
}
\makeatother
\begin{document}

\title{Recurrent Modular Structure in Spiking Neural Network Reinforcement Learning\\

}

\author{\IEEEauthorblockN{Vikram Ramavarapu}
\IEEEauthorblockA{\textit{Siebel School of Computing} \\
\textit{and Data Science} \\
\textit{University of Illinois} \\
\textit{Urbana-Champaign}\\
Champaign, IL, USA \\
vikramr2@illinois.edu}
\and
\IEEEauthorblockN{Himanshu Udupi}
\IEEEauthorblockA{\textit{Siebel School of Computing} \\
\textit{and Data Science} \\
\textit{University of Illinois} \\
\textit{Urbana-Champaign}\\
Champaign, IL, USA \\
hudupi2@illinois.edu}
\and
\IEEEauthorblockN{Ashish Gautam}
\IEEEauthorblockA{\textit{Computer Science and} \\
\textit{Mathematics Division} \\
\textit{Oak Ridge National Laboratory}\\
Oak Ridge, TN, USA \\
gautama@ornl.gov}
}

\IEEEaftertitletext{%
  \vspace{0.5\baselineskip}
  \centering
  \includegraphics[width=\textwidth]{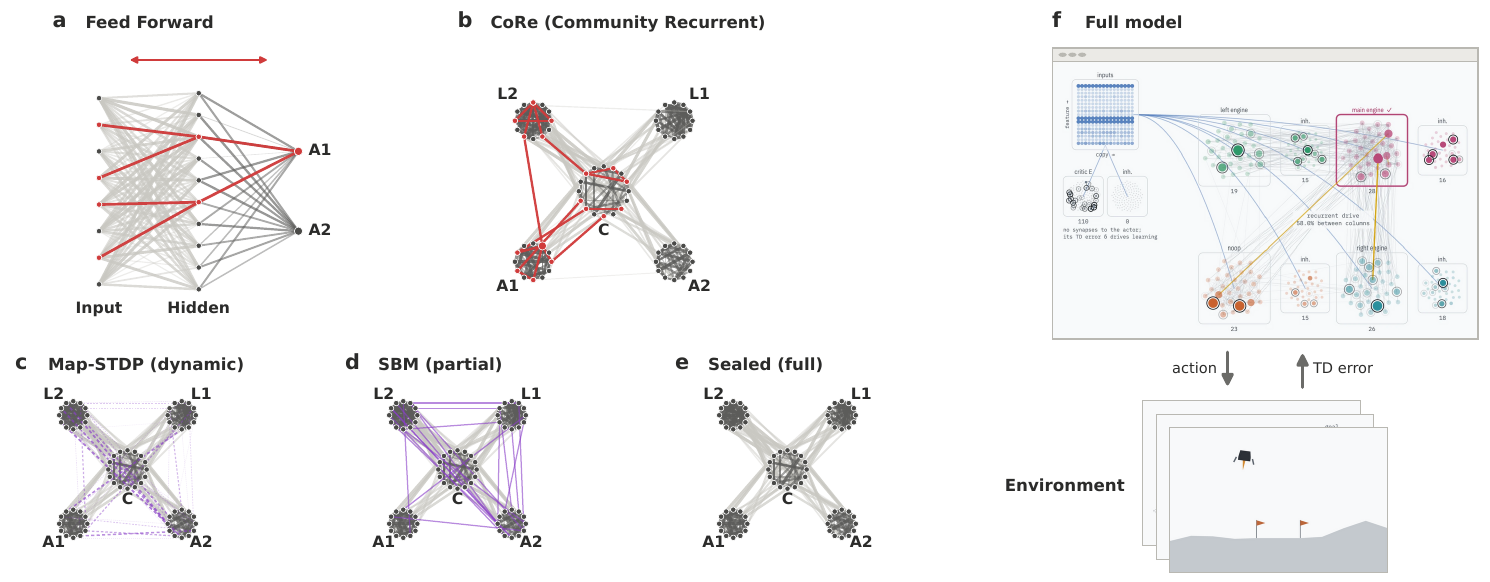}\\[4pt]
  \parbox{\textwidth}{\footnotesize\refstepcounter{figure}\label{fig:concept}%
    Fig.~\thefigure. \ Concept overview. (a)~A feed-forward network passes information layer by layer and is usually trained by backpropagation. Each action is a single output neuron. (b)~A community recurrent SNN (CoRe SNN) organises neurons into communities with distinct roles instead of layers. Each action is an entire community, and information flows through recurrent connections within and between communities. (c--e)~Three ways to obtain the communities: (c)~dynamic modularity, where Map-STDP encourages modular structure during learning, (d)~partial modularity, where a stochastic block model (SBM) initialization imposes it partially, and (e)~full modularity, where sealing imposes it fully. (f) The full model with CoRe inserted on a TD-STDP actor-critic backbone. An interactive visualization of the CoRe SNN variants is at \protect\textcolor{blue}{\protect\url{https://vikramramavarapu.org/CoRe}}.}%
  \vspace{0.5\baselineskip}
}

\maketitle

\begin{abstract}
Biological neural circuits are both modular and recurrent, with dense communities sparsely linked to one another. Spiking neural networks (SNNs) for reinforcement learning are usually feed-forward, and it is unclear how to obtain functional modules in a recurrent SNN or what they buy. In this work, we introduce a class of spiking networks, called Community Recurrent SNNs (CoRe SNNs): recurrent networks partitioned into communities with fixed roles. A controller community receives the stimulus, each action has its own community, and latent communities carry internal state. Spiking activity is then a walk between communities, and the action is the community the walk reaches. All learning is local and three-factor, and the recurrent communities are closer to cortical microcircuits than a feed-forward actor. We compare three regimes for obtaining the communities: dynamic modularity, in which a new learning rule we introduce, Map-STDP, encourages modular structure during learning; partial modularity, in which a stochastic block model (SBM) initialization enforces it partly; and full modularity, in which sealing enforces it fully. On CartPole, across three spiking neuron models, Map-STDP improves learning over the same network without it (+65 to +87 mean return over training), matches full modularity (sealed communities) on two of the three models, and makes the observed spike walk more compressible. Replacing the feed-forward actor of a TD-STDP actor-critic with recurrent action communities improves on the feed-forward network on CartPole (+25 to +45) and LunarLander (+36 to +41 final return), and matches it on Acrobot. How strongly to enforce modularity depends on the task: full modularity helps most on Acrobot, while on LunarLander the dynamic, partial and full regimes perform alike. Overall, these recurrent modular networks can match or even improve upon feed-forward approaches, with improvements in bio-plausibility and energy efficiency.
\end{abstract}


\section{Introduction}

The cortex is organized into columns~\cite{mountcastle1997columnar}, and its wiring forms densely connected communities that are sparsely linked to one another~\cite{meunier2010modular,sporns2016modular}. Modularity lets a circuit divide its work among parts that can be specialized and reused. Spiking neural networks (SNNs) are attractive models of such circuits and, on neuromorphic hardware, an energy-efficient way to run them~\cite{merolla2014million,davies2018loihi}. Yet SNNs for reinforcement learning (RL) are usually feed-forward stacks of layers, trained by backpropagation or by local reward-modulated rules~\cite{chung2020}. It is unclear how a recurrent SNN could acquire functional modules using only local plasticity, or what those modules would buy.

A guiding principle comes from compression. Hinton and van Camp argued that a good network is a simple one and trained networks by minimizing the description length of their weights~\cite{hinton1993keeping}. Efficient-coding theories make a similar claim for neural activity: neural systems should transmit information at low cost~\cite{barlow1961possible}, and the same accuracy--complexity trade-off underlies the free-energy principle~\cite{friston2010free}. As such, the map equation gives the description length of a random walk on a graph split into communities~\cite{rosvall2008maps}, and it is shortest when the communities contain the walk. Spiking activity is a walk over synapses, so a network whose activity is compressible has activity confined within modules and sparse traffic between them.

This view recalls the self-organizing map (SOM)~\cite{kohonen1982,kohonen1990}, the classic example of structure that emerges from local, unsupervised learning. Both approaches replace a global error signal with a local rule whose collective effect is organization. They differ in what is organized and why. A SOM arranges a layer of prototype units on a fixed lattice so that it preserves the topology of the input. Our network organizes its own recurrent connectivity into communities with distinct roles, using spikes, a reward-modulated rule, and the description length of its activity as the organizing objective. We see this as a modern descendant of the SOM idea, brought to recurrent SNNs, reinforcement learning, and information-theoretic compressibility.

We introduce Community Recurrent SNNs (CoRe SNNs), a class of recurrent networks partitioned into communities with fixed roles: a controller community that receives the stimulus, one community per action, and latent communities that carry internal state. We make four contributions.
\begin{itemize}
  \item \textbf{CoRe SNNs.} Spiking activity is a walk between communities, and the action is the community the walk reaches.
  \item \textbf{Map-STDP.} A local, three-factor learning rule that encourages the communities, and lowers description length during learning. 
  \item \textbf{Modularity regimes.} A comparison of dynamic, partial and full modularity across four neuron engines.
  \item \textbf{A drop-in actor.} CoRe SNNs replace the feed-forward actor of a TD-STDP actor--critic and are evaluated on CartPole, Acrobot and LunarLander, together with a sweep of cross-module connectivity.
\end{itemize}

\section{Background}

\subsection{Spiking Neural Networks}

SNNs communicate through discrete spike events, enabling sparse, temporally-coded computation that maps naturally to neuromorphic hardware~\cite{malcolm2023}. This makes them a closer model of cortical computation than conventional artificial neural networks. Because computation is event-driven, a neuron consumes energy mainly when it spikes or receives a spike, and can therefore run at a fraction of the energy of dense matrix multiplication on digital neuromorphic chips~\cite{merolla2014million,davies2018loihi} and on analog memristive crossbars. In a crossbar, each synapse is a conductance, and multiply--accumulate happens in place~\cite{prezioso2015training}. These savings depend on activity staying sparse~\cite{davies2021advancing}. Training SNNs is difficult because the spike function is non-differentiable. Surrogate-gradient backpropagation through time (BPTT)~\cite{neftci2019} works well but is non-local, needing weight transport and the full activity history, which have no known biological counterpart~\cite{lillicrap2020backpropagation} and are hard to implement with the local plasticity that neuromorphic hardware supports, so such networks are typically trained offline and then deployed~\cite{neftci2019,davies2018loihi}.

Spike-timing-dependent plasticity (STDP) is a local alternative to BPTT. A synapse is potentiated when a presynaptic spike shortly precedes a postsynaptic spike and depressed when the order is reversed~\cite{markram1997regulation,bi1998synaptic}. The update depends only on quantities available at the synapse, and the pairing of pulses across a device maps directly onto memristive hardware~\cite{prezioso2015training}. STDP alone is unsupervised, however. STDP alone is unsupervised: it strengthens whatever correlations the input contains. In networks that respect Dale's law, where each neuron is either excitatory or inhibitory, it can still learn useful features, for example digit prototypes shaped by lateral inhibition~\cite{diehl2015unsupervised}. It cannot, however, tell behaviorally useful correlations from irrelevant ones, so on its own it cannot learn a task defined by reward.

Three-factor rules add a global modulatory signal to the two local factors of STDP~\cite{fremaux2016neuromodulated,gerstner2018eligibility}. Rewards arrive seconds after the spikes that caused them, far outside the tens-of-milliseconds STDP window. Each synapse therefore keeps an eligibility trace $e_{ij}$. The trace accumulates pre--post coincidences and decays with a time constant matched to the delay of the outcome. A broadcast third factor $M(t)$ then turns the trace into a weight change:

$$
\dot{e}_{ij} = -\frac{e_{ij}}{\tau_e} + \mathrm{STDP}_{ij}(t),
\qquad
\Delta W_{ij} \propto M(t) e_{ij}(t).
$$

Chung and Kozma~\cite{chung2020} introduced a three-factor rule in which the temporal-difference (TD) reward prediction error $\delta = r + \gamma V(s') - V(s)$~\cite{schultz1997neural} is the third factor. In a spiking actor--critic network, this TD-modulated STDP was applied to standard control tasks such as CartPole and LunarLander.

\subsection{Neuron Models}

We use three spiking neuron models, which we call engines. In every engine, a
spike of presynaptic neuron $j$ gives postsynaptic neuron $i$ the
output-normalized kick
\begin{equation}
  b_{ij} = \mathrm{sgn}_j \, w_0 \, \frac{W_{ij}}{d_j}, \qquad
  d_j = \sum_k W_{kj}.
\end{equation}

\subsubsection{Branching Neuron Model}

The branching model is the standard neuron model of the SuperNeuroMAT
simulator~\cite{date2026superneuromat} and follows the branching-process
description of cortical activity~\cite{beggs2003}. The membrane forgets after
every step, so one step is one hop of the walk:
\begin{equation}
  \begin{gathered}
  u_i(t+1) = \sum_j b_{ij} x_j(t) + 2\theta \mu_i(t), \\
  x_i(t+1) \sim \mathrm{Bernoulli}\left( \mathrm{clip}\left(
    \frac{u_i(t+1)}{2\theta}, 0, 1 \right) \right),
  \end{gathered}
\end{equation}
where $x_i(t) \in \lbrace 0, 1 \rbrace$ is the spike indicator, $\theta$ the
threshold and $\mu_i$ the external (stimulus) drive.

\subsubsection{Leaky Integrate-and-Fire (LIF)}

The LIF neuron~\cite{lapicque1907} is driven by background noise as in
Brunel~\cite{brunel2000}:
\begin{equation}
  \begin{split}
  \tau_m \frac{dV_i}{dt} &= -(V_i - E_L) + \mu_i(t)
    + \sigma_V \sqrt{2 \tau_m}\, \xi_i(t) \\
    &\quad + \tau_m \sum_j b_{ij} \sum_f \delta\left( t - t_j^f \right),
  \end{split}
\end{equation}
with a spike and reset to $V_r$ when $V_i$ reaches $V_{th}$. Here $\xi_i$ is
unit white noise standing for background input, and $t_j^f$ are the spike
times of $j$.

\subsubsection{Quadratic Integrate-and-Fire (QIF)}

The QIF neuron~\cite{ermentrout1986} is written in the normalization of
Montbri\'{o} et al.~\cite{montbrio2015}:
\begin{equation}
  \begin{split}
  \tau \frac{dV_i}{dt} &= V_i^2 + \eta_i + I_i^{ext}(t)
    + \sigma \sqrt{\tau}\, \xi_i(t) \\
    &\quad + \tau \sum_j b_{ij} \sum_f \delta\left( t - t_j^f \right),
  \end{split}
\end{equation}
with a spike at $V_i = +V_{peak}$ and reset to $-V_{peak}$. Here $\eta_i$ is
the excitability.

\subsection{Random Walks, Information Theory, and Communities}

A random walk on a weighted directed graph is a process in which a walker at node $j$ moves to node $i$ with probability proportional to the weight of the edge $j \to i$. It is a Markov chain with transition probabilities $T_{ij} = W_{ij} / \sum_k W_{kj}$, and over time the fraction of steps spent at each node converges to a stationary distribution $\pi$ that measures how much flow each node carries. This is the idea behind PageRank~\cite{pagerank}. In a graph with communities, a walker that has entered a densely connected group tends to stay there for many steps before crossing a sparse link, so the walk's behavior reveals the community structure. In a spiking network, a natural walk follows spikes forward from each neuron to the neurons it excites.

Info-theoretic entropy can be measured by the description length of a random walk, following the core principle of information theory: the more predictable a process, the shorter its description. The map equation~\cite{rosvall2009map} measures the description length of a random walk on a graph partitioned into communities:

\begin{equation}
  L(M) = q_{\curvearrowright} H(\mathcal{Q}) + \sum_{i=1}^{m} p_{\circlearrowright}^i H(\mathcal{P}^i).
  \label{eq:map-equation}
\end{equation}

Each step of the walk is given a binary codeword, shorter for more frequent steps, and the entropy of a codebook is its average codeword length in bits. The map equation uses a two-level code. An index codebook $\mathcal{Q}$ names the module entered at each exit, which happens with probability $q_{\curvearrowright}$ per step. Each module $i$ has its own codebook $\mathcal{P}^i$ for naming the nodes inside it, used at rate $p_{\circlearrowright}^i$. Leaving a community costs extra bits, so the map equation is shorter for partitions that capture the graph's community structure. The partition that minimizes $L(M)$ is the one that best compresses the walk and reveals the communities, hence making the map equation the objective function of the community detection algorithm, Infomap~\cite{rosvall2008maps}. The map equation has been used to detect communities in networks from social~\cite{rosvall2009map}, biological~\cite{rosvall2010mapping}, and to study the modularity of cortical networks~\cite{sporns2016modular}.

A basic model that generates networks with community structure is the stochastic block model (SBM)~\cite{holland1983stochastic}. In an SBM, each node belongs to one of $K$ communities, and the probability of an edge between two nodes depends on a $K \times K$ matrix of connection probabilities. The SBM can generate networks with strong communities, weak communities, or no communities at all, depending on the choice of the probability matrix.

\subsection{Other Graph-Theoretic Approaches to SNNs}
Understanding how the structure of a network shapes its dynamics in the brain has been a long standing question in the neuroscience community \cite{curto2019relating}. For threshold-linear rate networks it has a concrete answer. A CTLN takes a directed graph, assigns the weights from it, and its fixed point supports, stable states and dynamic attractors are then determined by the graph \cite{curto2019fixed,curto2023graphrules,curto2024stable,parmelee2022coremotifs}. Brief external kicks switch the network from one attractor to the next, and this has been enough to assemble counters, sequences and gait controllers by hand \cite{parmelee2022sequential,londonoalvarez2026attractor}. Our SNN work sits next to this. We ask the same thing of a spiking network, and in both settings the answer takes the same shape, a collection of attractors pinned down by connectivity plus a set of input-triggered jumps between them, which is also how a finite state machine lives inside a recurrent network \cite{omlin1996constructing}.

Part of the CTLN picture has been shown to survive the move to spikes, but only in a restricted setting. In clustered E-I networks of Hawkes neurons the excitatory rates obey a CTLN in the mean-field limit, and the CTLN fixed points become metastable states of the finite network \cite{lienkaemper2025hawkes}. CTLNs also come with a modular story, where the attractors of a graph assembled from smaller pieces are inherited from the pieces \cite{egassantander2022nerve,londonoalvarez2026lowrank}, though no such result yet exists for spiking networks. Whether either property carries over to the SNNs we study is open. If it does, it would suggest a use for these rules in learning. A reward machine turns a history-dependent RL task into one with a finite set of states \cite{toroicarte2022jair}, and if those states could be planted as the attractors of a glued graph, a spiking network started from that graph would have the states in place and would only need to learn the jumps \cite{rigotti2010internal}. Some known initialisations of structure are detailed below. 

Columnar organization has also been used for supervised learning in spiking networks. CoLaNET~\cite{kiselev2025} gives each target class its own column, built from several layers of neurons and sub-class microcolumns, and trains it with local anti-Hebbian and dopamine-modulated plasticity. It was evaluated on classifying MNIST images~\cite{colanet2024} and Pong states by their proximity to reward, a task derived from model-based reinforcement learning. Its signal path is feed-forward, with only local winner-take-all competition and gating inside each column. The columns are fixed by design as class-specific classifiers, and the network evaluates states without selecting actions. In contrast, our work learns recurrent modular structure with Map-STDP and studies the effect of modularity on reinforcement learning.

\section{Map-STDP Learning Rule}
\label{sec:derivation}

Community detection solves a \emph{forward} problem: given a graph, find the partition into communities that minimizes the description length of a random walk on it, the map equation~\cite{rosvall2008maps}. Map-STDP solves the \emph{inverse} problem. The partition is fixed in advance (one controller module, one module per action, and a few latent modules), and plasticity shapes the synapses until that partition is the minimum-description-length one. This section states the rule and what each part does. The derivation from the map equation is in Appendix~\ref{app:derivation}.

\subsection{Ingredients}

Each synapse from neuron $j$ to neuron $i$ has a magnitude $W_{ij} \ge 0$ and the sign of its presynaptic type, $\mathrm{sgn}_j = \pm 1$. Each neuron belongs to one module $m(j)$: the controller $\mathcal{C}$, an action module in $\mathcal{A}$, or a latent module in $\mathcal{L}$.

\begin{itemize}
  \item \textbf{The walk.} A step of the walk from $j$ to $i$ is a spike of $i$ within a short window after a spike of $j$. Its flow $J_{ij}$, coarse-grained to modules, is a small Markov chain over the network's own states. The policy is the flow into the action modules, $P(a \mid \mathbf{o}_f) \propto p_a(f)^\beta$, where $\mathbf{o}_f$ is the stimulus at frame $f$, $p_a(f)$ is the flow into action module $a$ in that frame, and $\beta$ is a temperature parameter.
  \item \textbf{Description length.} $D(W) = \mathbb{E}_f[L(J_f)]$ is the code length of the walk averaged over frames. Leaving module $m$ costs $M^{\ast}_m = \partial L / \partial q_m \ge 0$ bits per unit of exit flow $q_m$. It is computed from $K$ module-level statistics and broadcast to module $m$: it is the rule's neuromodulator.
  \item \textbf{Signed counter.} $c_{ij}$ is the change in $i$'s spikes, within the window, caused by $j$'s kicks alone. It is the local pre-synaptic activity times post-synaptic sensitivity, and carries $\mathrm{sgn}_j$.
\end{itemize}

\subsection{The rule}

The per-frame update of each synapse is
\begin{equation}
  \begin{split}
  \Delta W_{ij}
  &= \underbrace{\eta \delta_f e_{ij}}_{\text{task}} \\
  &\quad - \underbrace{\lambda M^{\ast}_{m(j)}
      \left( \mathbb{I}[i \notin m(j)] - \bar{e}^{obs}_j \right)
      c_{ij}}_{\text{structure}} \\
  &\quad - \underbrace{\epsilon_h (d_j - 1) \frac{\pi_j W_{ij}}{d_j}}_{\text{homeostasis}}.
  \end{split}
  \label{eq:full}
\end{equation}
The rule has three terms:
\begin{itemize}
  \item \textbf{Task.} $\delta_f$ is a global TD error and $e_{ij}$ an
    action-gated eligibility trace built from $c_{ij}$ (Eq.~\eqref{eq:task}).
  \item \textbf{Structure (the map term).} $\bar{e}^{obs}_j$ is the share of
    $j$'s observed transitions that leave its module. Transitions that leave
    the module are pushed down when the module leaks, and those that stay are
    pushed up, scaled by $M^{\ast}_{m(j)}$.
  \item \textbf{Homeostasis.} $d_j$ is $j$'s total output weight, held near 1.
    It rescales a column uniformly, so it does not change the walk.
\end{itemize}
$\lambda$ sets the strength of the map term. In partial and full modularity, $\lambda = 0$ and the map term is not applied. In dynamic modularity, $\lambda > 0$ and the map term is applied. Inhibitory synapses reverse the signs of the structural term, through $\mathrm{sgn}_j$ in $c_{ij}$. On average the structural term is gradient descent on $D$, up to a rescaling that cannot change $D$ (Appendix~\ref{app:proof}). The controller's synapses onto the action and latent modules carry the policy, so they are exempt from the map term (Appendix~\ref{app:routing}).

\subsection{Three regimes of modularity}

The experiments compare three regimes:
\begin{itemize}
  \item \emph{partial} modularity: $\lambda = 0$ on a network initialized with SBM connectivity ($p_{cross} = 0.05$), so communities are only partly enforced and the task term alone shapes the structure;
  \item \emph{full} modularity: $\lambda = 0$ on a network built without cross-module synapses ($p_{cross} = 0$, sealed);
  \item \emph{dynamic} modularity: the complete rule~\eqref{eq:full}, which encourages the communities during learning.
\end{itemize}

\section{Model Architecture}

\subsection{Standalone Map-STDP Agent}

The controller has no recurrent synapses of its own. In each environment frame the network is simulated on the current weights:
\begin{itemize}
  \item 100 steps for the branching engine;
  \item 100~ms for the LIF and QIF engines;
  \item 20 hops of power iteration for the flow engine, which has no spikes.
\end{itemize}
The spike raster gives the module flow (spike shares), the pair counts and the signed counters. Each spiking agent runs 8 parallel heads on shared weights and pools their statistics.

\subsection{TD-STDP Actor-Critic Backbone}

The backbone is the TD-STDP actor--critic of Chung and Kozma. Its neurons are noisy LIF ($\tau_m = 20$~ms) on a 1~ms grid, and each environment step is a 20~ms window. Population sizes follow the authors' configurations on CartPole, and in Acrobot the critic has 40 excitatory neurons and the actor has 20 per action; on LunarLander each excitatory group is paired with an inhibitory group of equal size (128 critic, 32 per action). CoRe replaces the feed-forward actor with recurrent communities, one per action, among the actor's excitatory neurons. Actions are read out by a race between the action communities' excitatory spike counts instead of the original softmax.

\subsection{SBM Initialization}

We initialize the network with an SBM with $K=|\mathcal{C}| + |\mathcal{A}| + |\mathcal{L}|$ communities. The intra-community connection probability, $p_{in}$ is consistently set to 0.5. In the sealed regime, the inter-community connection probability, $p_{cross}=0$ with all-to-all connections only between the controller and the other communities. In most experiments, the partial regime has $p_{cross}=0.05$, but in the inter-module connectivity study, we vary $p_{cross}$ from 0 (full modularity) to 0.5 (no communities).

\section{Implementation Details}

Models are implemented in NumPy, with Numba kernels for the LIF and QIF engines. The branching engine uses SuperNeuroMAT's~\cite{date2026superneuromat} branching mode (infinite leak, uniform threshold noise), so each simulation step is one hop of the walk. Environments come from Gymnasium~\cite{towers2024gymnasium}.

\section{Training}

\paragraph{Standalone agent.}
Agents are trained on CartPole for 1500 episodes, with 20 seeds per condition. Each condition is one engine (flow, branching, LIF, QIF) in one of three regimes: dynamic ($\lambda = 0.05$), partial ($\lambda = 0$), or full (sealed: $\lambda = 0$, no cross-module synapses). Map-STDP and the task update are applied once per frame. All conditions use TD-error normalization and an adaptive readout sharpness. In the spiking engines, $\lambda$ and the task rate are rescaled to match the flow engine's update size at initialization.

\paragraph{Backbone.}
The TD-STDP and Map-STDP updates are applied once per 20~ms window, without Adam. Input scales are calibrated so that every arm starts at the same 40~Hz rate. The arms are the original feed-forward network, the feed-forward network with the race readout, and CoRe SNNs in the partial, dynamic and full regimes. Each arm has 20 seeds. Per task:
\begin{itemize}
  \item \textbf{CartPole:} 400 episodes, one environment copy, race bound 16.
  \item \textbf{Acrobot:} 500 episodes, one environment copy, race bound 16,
        one action every 40~ms, plus a firing-rate cost and floor on the
        actor.
  \item \textbf{LunarLander:} 10{,}000 episodes pooled over 16 environment
        copies on shared weights, race bound 26, one action every 40~ms,
        plus the same rate cost and floor.
\end{itemize}

\section{Results}
\label{sec:results}
\subsection{Minimal spiking policy network}

\subsubsection{Results on CartPole}
\begin{figure*}[t]
\centering
\includegraphics[width=\textwidth]{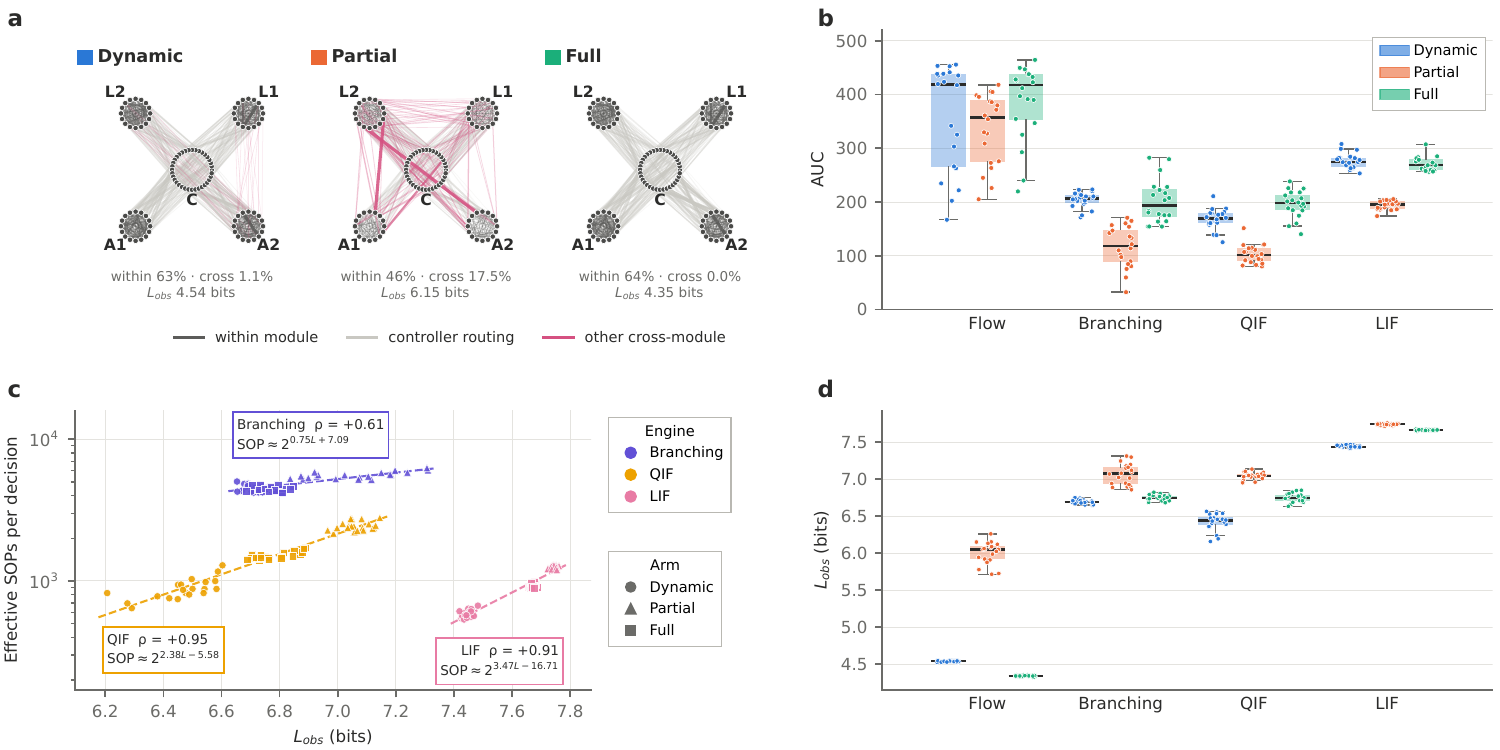}
\caption{(a) Trained CartPole networks on the flow engine, one seed (2000), from the same initial network. C is the controller, L1--L2 the latent modules and A1--A2 the action modules. Edges of at least 2\% of the maximum weight are drawn, with width and opacity proportional to weight. Below each network: the share of weight within modules and across modules (excluding controller routing), and $L_W$, the map equation (bits) of W's own walk. (b) AUC, the mean return over 1500 training episodes, for each engine and regime; 20 seeds per condition. (c) Effective synaptic operations per decision (spikes $\times$ outgoing synapses above the weight floor, averaged over 50 held-out stimuli with W frozen) against $L_{obs}$, for the final networks of the three spiking engines. Color: engine; shape: regime. Dashed lines are least-squares fits of $\log_{10}$ effective SOPs on $L_{obs}$; $\rho$ is Spearman's. Post-hoc, exploratory analysis. (d) $L_{obs}$, the held-out description length (bits) of the engine's observed walk, at episode 1500. In (b) and (d), boxes show the median and interquartile range, whiskers the full range, and dots individual runs.}
\label{fig:emergent-v4}
\end{figure*}

Figure~\ref{fig:emergent-v4} summarizes the comparison. In panel (a) of the figure, the trained networks of one seed are shown for each regime of the flow engine with a central controller, two latent modules, and two action modules. In the dynamic and full regimes, the modules are clearly apparent, indicating that qualitatively, on this engine, the map term is effective in shaping the network into a modular structure. In the partial regime, the modules are less distinct, indicating that the SBM initialization alone leaves room for inter-module connectivity. In panel (b), the mean return over training is shown for each engine and regime. The dynamic and full regimes outperform the partial regime across all engines, however overall, flow outperforms the other engines, albeit with a larger spread. LIF comes next, followed by QIF and branching. This may be due to the fact that in the base CoRe architecture, the random walk entropy formulation is most closely aligned with the flow engine, and spiking engines approximate walk behavior. In panel (c), the effective synaptic operations per decision (spikes $\times$ outgoing synapses above the weight floor, averaged over 50 held-out stimuli with W frozen) is plotted against $L_{obs}$, the held-out description length of the observed walk. As there is a direct exponential correlation, this may hint at the fact that the more compressible the walk, the fewer synaptic operations are needed to make a decision, and therefore the more energy efficient the network is. In panel (d), $L_{obs}$ is shown for each engine and regime at episode 1500. The dynamic and full regimes have lower $L_{obs}$ than the partial arm, indicating that the modularity is effective in shaping the network into a more compressible structure. Compressibility largely follows from modularity itself: the full regime, which has no map term, has a lower $L_{obs}$ than the partial regime on all four engines, and on flow and branching it is within 0.2 bits of the dynamic regime (Table~\ref{tab:emergent-v4}). On QIF and LIF the map term lowers $L_{obs}$ further. Overall, these results suggest that the map term is effective in shaping the network into a modular structure that is more compressible and energy efficient, and that this leads to improved performance on the CartPole task.

\subsubsection{Latent Ablation}

To find out what the latent modules contribute, we ablated them in the final CartPole models (episode 1500) of every engine and regime (240 models in all) and measured the change in return with plasticity off. The design was pre-registered. Each network has a 36-neuron controller $\mathcal{C}$, two latent modules $\mathcal{L} = \mathcal{L}_1 \cup \mathcal{L}_2$ and two 16-neuron action modules $\mathcal{A}$. Because every engine normalizes outgoing weights, $T_{ij} = W_{ij} / d_j$ with $d_j = \sum_k W_{kj}$, a cut either loses the drive the cut synapses carried or reroutes it, so we test both cases. A \emph{lesion} cuts the latents' outgoing synapses and leaves every other neuron's $d_j$ unchanged, so the drive into the latents is lost. It tests whether the latents \emph{relay} anything the policy uses. A \emph{removal} disconnects the latents entirely, and each remaining neuron's outflow renormalizes onto its other targets, $T'_{ij} = T_{ij} / (1 - \ell_j)$ with $\ell_j = \sum_{k \in \mathcal{L}} T_{kj}$. It tests whether the latents act as a \emph{sink} that sets the action modules' share of the controller's drive. In the \emph{latent path only} condition the direct controller-to-action synapses are cut, which tests whether the latents alone can carry the policy. Each model is evaluated intact and under each ablation for 50 episodes with common random numbers, so every model is its own control. QIF's warm state is not saved, so its intact returns differ slightly from training.

The policy is carried by the direct routing, not by the latents. With it cut, every group fell to about random play (returns of 10--32 against 22.8 for a random policy; Table~\ref{tab:ablation}, last column). Removing the latents lowered return most in flow ($-206$ to $-287$), branching in the dynamic regime ($-118$) and LIF in the dynamic and full regimes ($-112$, $-80$), and by $-31$ to $-65$ in the other groups. Without the latents the controller's whole outflow reaches the action modules, and their share of the drive rose from 0.08--0.49 to 0.57--0.80. The latents therefore act as a sink that sets how much of the controller's drive reaches the actions, and the trained policy depends on that setting.

Their output is not a relay the policy needs. In full-modularity models the lesion left returns identical to intact, since they have no latent-to-action synapses, and in the dynamic regime it changed almost nothing on QIF and LIF. Where the latents still drive the actions, cutting their output \emph{improved} return (branching dynamic $+47$, LIF partial $+38$), so it acted as interference. Flow in the dynamic regime is the exception: lesioning one latent module cost return ($\mathcal{L}_2$: $-117$, $\mathcal{L}_1$: $-77$), but lesioning both did not ($-3$), which suggests that the two modules' small outputs balance each other. Differences between regimes were small. Only LIF (partial against dynamic) and branching (full against dynamic) differed, and the remaining contrasts did not.

\begin{table}[t]
\centering
\footnotesize
\setlength{\tabcolsep}{2pt}
\renewcommand{\arraystretch}{1.15}
\caption{Change in return from the intact baseline when the latents are
ablated (median over the 20 models in each row). $^{\dagger}$Identical to
intact on every model. Intact: mean return of the intact models. 
}
\label{tab:ablation}
\begin{tabular}{@{}llrrrrrr@{}}
\toprule
 & & & \multicolumn{3}{c}{Lesion} & Removal & Latent \\
\cmidrule(lr){4-6}
Engine & Regime & Intact & $\mathcal{L}$ & $\mathcal{L}_1$ & $\mathcal{L}_2$ & $\mathcal{L}$ & path only \\
\midrule
\multirow{3}{*}{Flow} & dynamic & $462$ & $-3$ & $-77$ & $-117$ & $-206$ & $-455$ \\
 & partial & $414$ & $-18$ & $+1$ & $0$ & $-287$ & $-418$ \\
 & full & $481$ & \same & \same & \same & $-254$ & $-478$ \\
\midrule
\multirow{3}{*}{Branching} & dynamic & $199$ & $+47$ & $+18$ & $+16$ & $-118$ & $-167$ \\
 & partial & $124$ & $+18$ & $+12$ & $+8$ & $-65$ & $-105$ \\
 & full & $178$ & \same & \same & \same & $-41$ & $-137$ \\
\midrule
\multirow{3}{*}{QIF} & dynamic & $150$ & $0$ & $+1$ & $+1$ & $-31$ & $-140$ \\
 & partial & $96$ & $+13$ & $+6$ & $+13$ & $-50$ & $-87$ \\
 & full & $192$ & \same & \same & \same & $-32$ & $-179$ \\
\midrule
\multirow{3}{*}{LIF} & dynamic & $327$ & $+2$ & $+1$ & $-2$ & $-112$ & $-311$ \\
 & partial & $199$ & $+38$ & $+18$ & $+17$ & $-42$ & $-181$ \\
 & full & $305$ & \same & \same & \same & $-80$ & $-281$ \\
\bottomrule
\end{tabular}
\end{table}

When proceeding with the TD-STDP backbone, we test models without latents.

\subsection{TD-STDP actor-critic backbone}

\subsubsection{Performance}

\begin{figure*}[t]
\centering
\includegraphics[width=\textwidth]{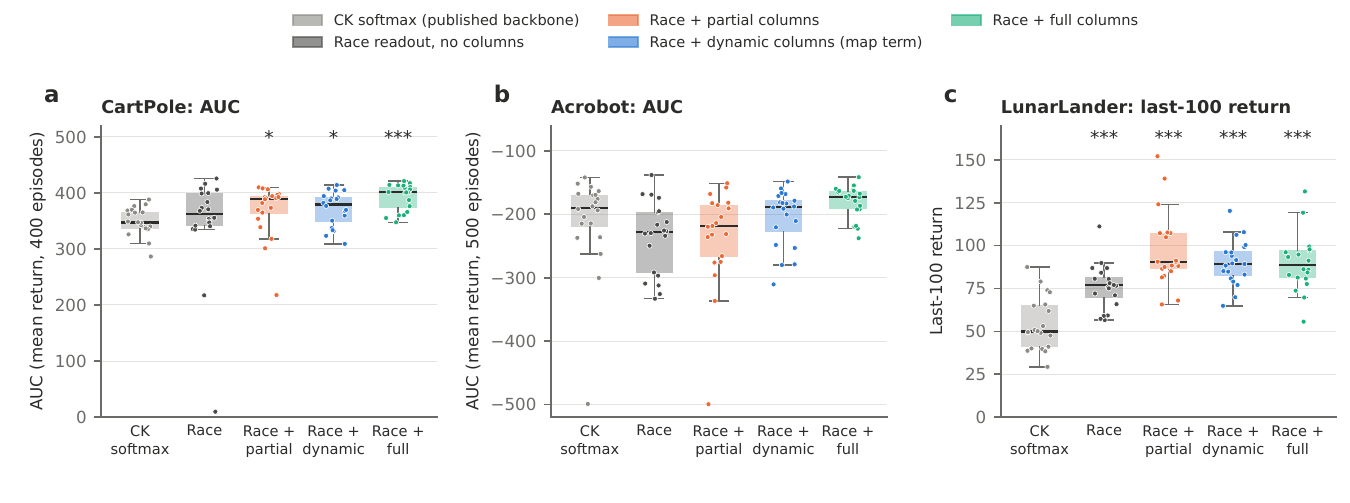}
\caption{TD-STDP actor-critic backbone. The actor network generates actions based on the current state, while the critic network evaluates the actions. The TD-STDP learning rule updates the weights based on the prediction error and the eligibility trace. In panels (a-c), the mean return over training episodes is shown for each environment and arm; 20 seeds per arm. Lunar Lander shows mean return for the last 100 episodes.}
\label{fig:td-stdp-backbone}
\end{figure*}

In the previous subsection, we showed that the minimal spiking policy network can be significantly improved with CoRe. However, in order to show that advancements can be made with this abstraction, we now demonstrate how CoRe can be applied to an existing TD-STDP actor-critic backbone to improve its performance. Figure~\ref{fig:td-stdp-backbone} shows the mean return over training episodes for each environment and arm. The results indicate that when CoRe is applied to the TD-STDP backbone, the performance of the network improves cleanly in CartPole and LunarLander. However, the results on Acrobot are a bit more complicated, when the race readout first replaced the softmax, the performance dropped, but when CoRe was applied, the performance improved and slightly exceeded the original softmax performance in the full regime. Overall, these results suggest that CoRe can be effectively applied to existing TD-STDP actor-critic backbones to improve their performance on reinforcement learning tasks.

\subsubsection{Inter-module Connectivity Effects}

\begin{figure}[t]
\centering
\includegraphics[width=0.4\textwidth]{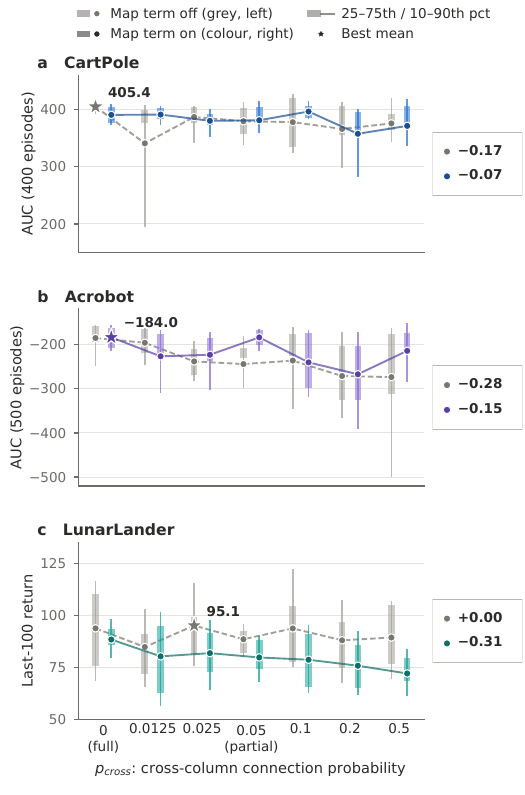}
\caption{Inter-module connectivity effects in the TD-STDP actor-critic backbone. The figure illustrates how the connectivity between different modules influences the performance and learning dynamics of the network. The Spearman coefficients are shown to the right of each plot, indicating the strength and direction of the relationship between inter-module connectivity and performance.}
\label{fig:inter-module-connectivity-effects}
\end{figure}

In the previous sections, we have explored comparing the three regimes of modularity: partial (SBM with $p_{cross}=0.05$), full (sealed, $p_{cross}=0$), and dynamic (learned with the map term). However, it has been left open how much cross-module connectivity is optimal for performance, and what happens when $p_{cross}$ is varied. To address this, we conducted a sweep of $p_{cross}$ values from 0 (sealed) to 0.5 (no communities) in the TD-STDP actor-critic backbone, with both the map term on and off. When $p_{cross}=0$, while the map-term is constrained to act upon intra-module connectivity, this still yields a difference in performance compared to when the map term is off.

The results, shown in Figure~\ref{fig:inter-module-connectivity-effects}, indicate that optimal inter-module connectivity is task-dependent, and setting-dependent. Across all three environments, the most optimal $p_{cross}$ lies in its lower range. In CartPole, the performance is highest at $p_{cross}=0$ with the map term off. In Acrobot, the performance is highest at $p_{cross}=0$ with the map term on. In LunarLander, the performance is highest when $p_{cross}=0.025$ with the map term on. Overall, this suggests that in some tasks, actions may be dependent on each other, and therefore some cross-module connectivity is beneficial, while in other tasks, actions may be independent of each other, and therefore cross-module connectivity is detrimental.

The trends in the results are also task and setting dependent. While most results show a downward trend in performance as $p_{cross}$ increases, there are several exceptions and nuances. Many of the correlation coefficients are brought down by non-monotonicity in the trends, with many trends following a drop off, then a rise, then a drop again, and sometimes another rise. This is especially apparent with CartPole with the map term on, and LunarLander with the map term off.

Overall, these results suggest that the optimal inter-module connectivity is task-dependent, and setting-dependent, and that the map term can have a significant impact on performance. Future work should explore dynamically setting the $K \times K$ inter-module connectivity matrix, rather than a single $p_{cross}$ value, to allow for more flexibility in the network's structure and performance.

\section{Discussion: Biological Plausibility}

CoRe SNNs share two features of cortex. First, as motivated in the Introduction, their connectivity is modular: Map-STDP learns weights that are dense within communities and sparse between them, and partial and full modularity impose the same structure directly. Second, all CoRe models learn with three-factor rules, in which a local eligibility changes a weight only when a neuromodulatory signal arrives. Such rules have direct experimental support~\cite{gerstner2018eligibility,fremaux2016neuromodulated}, our task term follows the dopaminergic temporal-difference account of reward learning~\cite{schultz1997neural}, and the structural modulator is a single scalar per module, in line with the regional specificity of neuromodulation~\cite{engelhard2019specialized}. Two parts remain abstractions. The structural modulator is indexed by the presynaptic module, which requires a roughly uniform or projection-specific signal, and the per-neuron baseline $\bar{e}^{obs}_j$ has no established biological counterpart.

Map-STDP also fits the efficient-coding view of the Introduction. Communication accounts for much of the brain's energy budget~\cite{attwell2001energy}, and compressing the network's activity prunes cross-module traffic. In exploratory measurements, a lower description length coincided with fewer effective synaptic operations per decision (Spearman $\rho = 0.61$--$0.95$ across engines). The task term keeps the network useful for acting, and the structural term keeps its activity compressible. Finally, our TD-STDP actor-critic~\cite{chung2020} is more biologically plausible than the original. Its actors are recurrent communities, closer to cortical microcircuits and to the channel organization of the basal ganglia~\cite{douglas2004neuronal,redgrave1999basal}, it runs in a noisy leaky integrate-and-fire regime at lower firing rates, and it selects actions by a race between action communities, which is equivalent to softmax sampling under constant hazards~\cite{yellott1977}.

\section{Limitations and Future Work}

\textit{Limitations.} Map-STDP relies on a random-walk abstraction of inference. Random walks, and branching processes in particular, are inherently sequential, so estimating the walk statistics per frame becomes a bottleneck as networks scale; a more parallelizable abstraction of inference is needed. Second, the map equation is only one principle that can drive module formation. Alternatives such as the Constant Potts Model~\cite{traag2011}, modularity maximization~\cite{blondel2008}, attractor models~\cite{curto2018}, and the dynamic SBM matrices discussed earlier remain unexplored here. Our results suggest that such alternatives are worth testing, because compressibility largely accompanied modularity itself rather than requiring an explicit information-theoretic objective. Objectives that reward modularity directly could therefore yield similar structure without the map equation. Third, as Table~I shows, the latent communities act largely as sinks and, at best, routers; we do not yet train them to perform useful computation of their own.

\textit{CoRe as a starting point.} We view CoRe as a base architecture that opens many directions. Its modular structure decomposes functionality into smaller, separately trainable pieces, and the TD-STDP backbone experiment shows that a CoRe module can be inserted into and coupled with existing, performant architectures and improve them. This opens several directions. Latents could be trained to carry out subtasks or store memory, through persistent activity across frames or through coupling with memory-oriented architectures such as Hopfield networks~\cite{hopfield1982} or Boltzmann machines~\cite{ackley1985}. Coupling with architectures built for binding and unbinding~\cite{frady2020} in the controller could let different latents attend to different objects in the environment. Finally, the same modular interface could host other backbones and other module-forming objectives, so that CoRe serves as a common substrate for these extensions.

\section*{Acknowledgment}

We thank Dr. Ramakrishnan (Ramki) Kannan at Oak Ridge National Laboratory for providing access to his local cluster. We also thank the AI Interpretability @ Illinois group for providing the collaborative network to make this work possible.

The authors used Claude Code (Anthropic) with the Claude Sonnet 5.5, Opus 5.5, and Fable 5.1 models during this work. The AI generated software and experimental scripts used for the results in Section~\ref{sec:results}; assisted with the derivation in Appendix \ref{app:derivation}; assisted in the iterative design of the figures; and provided language editing throughout the manuscript. All code was reviewed, tested, and validated by the authors against the source data; all derivations were independently checked by the authors; and all text was reviewed and revised by the authors. The authors take full responsibility for the content of this paper, including the integrity of the analysis and results.

\appendices

\clearpage
\onecolumn
\raggedbottom

\section{Full Results}
\label{app:results}

\begin{table*}[!ht]
\centering
\small
\caption{CartPole, model v4 with the observed-walk rule: dynamic ($\lambda = 0.05$), partial ($\lambda = 0$) and full (sealed: no cross-module synapses, $\lambda = 0$) modularity, on four engines. Mean $\pm$ s.d.\ over 20 seeds (2000--2019), 1500 episodes per run. AUC: mean return over all episodes. Return: mean of the last 100 episodes. Frozen: mean return of 20 episodes with $W$ frozen at the end. $L_{obs}$: held-out description length (bits) of the engine's observed walk at episode 1500. Solved: runs whose trailing-100 mean reaches 475. $p$: two-sided Mann--Whitney, Holm-adjusted over the four engines, of each arm against the map-term-on arm of the same engine.}
\label{tab:emergent-v4}
\begin{tabular}{llccccccc}
\toprule
Engine & Regime & AUC & Return & Frozen & $L_{obs}$ (bits) & Solved & $p_{\mathrm{AUC}}$ & $p_{L_{obs}}$ \\
\midrule
Flow & Dynamic & 356.9 $\pm$ 96.7 & 442.4 $\pm$ 103.5 & 463.9 $\pm$ 94.3 & 4.54 $\pm$ 0.01 & 15/20 & -- & -- \\
 & Partial & 335.9 $\pm$ 64.9 & 431.6 $\pm$ 105.6 & 412.8 $\pm$ 122.8 & 6.00 $\pm$ 0.14 & 20/20 & 0.17 & $<10^{-6}$ \\
 & Full & 388.3 $\pm$ 68.8 & 475.9 $\pm$ 80.3 & 481.4 $\pm$ 80.1 & 4.34 $\pm$ 0.01 & 20/20 & 1.0 & $<10^{-6}$ \\
\midrule
Branching & Dynamic & 204.1 $\pm$ 14.1 & 213.2 $\pm$ 50.4 & 198.9 $\pm$ 90.2 & 6.69 $\pm$ 0.03 & 0/20 & -- & -- \\
 & Partial & 116.8 $\pm$ 37.5 & 128.7 $\pm$ 68.1 & 126.5 $\pm$ 108.8 & 7.06 $\pm$ 0.14 & 0/20 & $<10^{-6}$ & $<10^{-6}$ \\
 & Full & 201.4 $\pm$ 39.0 & 176.2 $\pm$ 67.9 & 177.5 $\pm$ 78.3 & 6.75 $\pm$ 0.04 & 3/20 & 1.0 & $5.9 \times 10^{-5}$ \\
\midrule
QIF & Dynamic & 168.3 $\pm$ 18.8 & 174.3 $\pm$ 39.7 & 151.6 $\pm$ 43.6 & 6.42 $\pm$ 0.11 & 0/20 & -- & -- \\
 & Partial & 103.4 $\pm$ 17.1 & 124.5 $\pm$ 91.1 & 97.5 $\pm$ 82.8 & 7.05 $\pm$ 0.05 & 0/20 & $<10^{-6}$ & $<10^{-6}$ \\
 & Full & 196.0 $\pm$ 24.3 & 198.0 $\pm$ 43.2 & 189.4 $\pm$ 30.7 & 6.74 $\pm$ 0.06 & 0/20 & 0.0023 & $<10^{-6}$ \\
\midrule
LIF & Dynamic & 276.2 $\pm$ 13.5 & 321.6 $\pm$ 80.8 & 310.2 $\pm$ 94.0 & 7.44 $\pm$ 0.02 & 0/20 & -- & -- \\
 & Partial & 194.0 $\pm$ 7.8 & 208.4 $\pm$ 40.4 & 199.7 $\pm$ 54.1 & 7.74 $\pm$ 0.01 & 0/20 & $<10^{-6}$ & $<10^{-6}$ \\
 & Full & 270.8 $\pm$ 12.7 & 306.5 $\pm$ 72.0 & 294.6 $\pm$ 80.6 & 7.66 $\pm$ 0.01 & 0/20 & 0.6 & $<10^{-6}$ \\
\bottomrule
\end{tabular}
\end{table*}

\section{Derivation of Map-STDP}
\label{app:derivation}

This appendix derives the rule in Section~\ref{sec:derivation}. All logarithms are base 2 (description lengths in bits).

\subsection{Setup and notation}

Each synapse from presynaptic neuron $j$ to postsynaptic neuron $i$ has a
magnitude $W_{ij} \ge 0$. The type of $j$ fixes its sign,
$\mathrm{sgn}_j = +1$ (excitatory) or $-1$ (inhibitory), as in Dale's law.
We write
\begin{equation}
  d_j = \sum_k W_{kj}, \qquad T_{ij} = \frac{W_{ij}}{d_j},
\end{equation}
for $j$'s total output weight and the share of it that goes to $i$. Each
neuron belongs to one module $m(j) \in \lbrace 1, \dots, K \rbrace$. The
module types are the controller $\mathcal{C}$, the action modules
$\mathcal{A}$ (one per action) and the latent modules $\mathcal{L}$.
Environment frames are indexed by $f$, and the stimulus of frame $f$ is
$\mathbf{o}_f$.

$T$ is column-stochastic: column $j$ says where $j$'s
output goes. A random walk that follows spikes forward moves from $j$ to $i$
with probability $T_{ij}$.

\subsection{The observed walk}
\label{app:walk}

A step of the walk from $j$ to $i$ is a spike of $i$ within a short window
$w$ after a spike of $j$. The expected number of such spikes splits into
three parts:
\begin{equation}
  \begin{split}
  E_{ij} &= \underbrace{a_i}_{\text{chance}}
         + \underbrace{D_{ij}}_{\text{direct}}
         + \underbrace{X_{ij}}_{\text{indirect}} \ge 0, \\
  a_i &= r_i w, \qquad D_{ij} = \mathrm{sgn}_j \tilde{\Lambda}_{ij}.
  \end{split}
  \label{eq:Eij}
\end{equation}
Here $r_i$ is $i$'s firing rate. $\tilde{\Lambda}_{ij} = h_i T_{ij}$ is the
direct effect of $j$'s kick on $i$, where $h_i$ is $i$'s windowed
sensitivity (gain times kick size). $X_{ij}$ collects chains $j \to k \to i$
and common input. The observed transition probabilities and link flows are
\begin{equation}
  E_j = \sum_i E_{ij}, \qquad
  T^{obs}_{ij} = \frac{E_{ij}}{E_j}, \qquad
  J_{ij} = \pi_j T^{obs}_{ij},
  \label{eq:Tobs}
\end{equation}
where $\pi_j$ is $j$'s share of all spikes in the frame.

\paragraph{Intuition.} $E_{ij}$ is the mean of a count, so it is
non-negative whatever the signs of the synapses. Inhibition does not create
negative transitions; it pushes $E_{ij}$ below chance.

\paragraph{Pair counts measure the walk.} With
$\kappa_{ij} = \sum_t x_j(t) N_i(t, t+w]$, where $x_j(t)$ marks a spike of
$j$ at step $t$ and $N_i$ counts spikes of $i$, stationarity within the
frame gives $\mathbb{E}[\kappa_{ij}] = \bar{n}_j E_{ij}$. So
\begin{equation}
  \hat{T}^{obs}_{ij} = \frac{\kappa_{ij}}{\sum_k \kappa_{kj}}, \qquad
  \hat{\bar{e}}^{obs}_j
    = \frac{\sum_{i \notin m(j)} \kappa_{ij}}{\sum_i \kappa_{ij}}.
  \label{eq:pair}
\end{equation}
$\bar{e}^{obs}_j$ is the share of $j$'s observed transitions that leave its
module. A neuron can estimate it from its own spike times and two population
signals: activity in its own module, and in the whole network.

\paragraph{The flow engine is the special case.} Without spikes there is no
chance ($a = 0$) and no indirect term ($X = 0$), and all gains are equal, so
$T^{obs} = T$. The stimulus $v(\mathbf{o})$ re-injects a share $\alpha$ of
the activity at each step:
\begin{equation}
  \pi = (1-\alpha) T \pi + \alpha v(\mathbf{o}), \qquad \sum_i \pi_i = 1.
  \label{eq:flow}
\end{equation}
For $\alpha > 0$ the solution is unique. Starting from $x_0 = v$, $n$
power-iteration hops give
$\lVert x_n - \pi \rVert_1 \le 2 (1-\alpha)^n$, whatever the network size.

\subsection{Communities as states and the policy}

Coarse-graining the frame's walk to modules gives a $K$-state Markov chain,
with tables indexed [post, pre]:
\begin{equation}
  p_a = \sum_{j \in a} \pi_j, \qquad
  J_{ba} = \sum_{i \in b} \sum_{j \in a} J_{ij}, \qquad
  T^K_{ba}(\mathbf{o}_f) = \frac{J_{ba}}{p_a}.
  \label{eq:TK}
\end{equation}
Because the stimulus is concentrated in the controller, the controller's
column of $T^K$ is the stimulus-dependent one. The action is read from the
flow into the action modules,
\begin{equation}
  P(a \mid \mathbf{o}_f)
    = \frac{p_a(f)^\beta}{\sum_{a' \in \mathcal{A}} p_{a'}(f)^\beta},
  \qquad a \in \mathcal{A},
  \label{eq:policy}
\end{equation}
with $\beta = 1$ the plain readout. Model v4 adapts $\beta$ to the trend of
episode return. In spiking engines the readout is a race between the action
modules' spikes, which samples~\eqref{eq:policy}.

\subsection{The map equation}
\label{app:deriv}

Describe the walk with a two-level code. Inside a module, name the next
neuron, or ``exit'', with a module-local codeword. On exit, name the next
module from a global index. The average code length per step
is~\cite{rosvall2008maps}
\begin{equation}
  \begin{split}
  L &= q_{\curvearrowright} \log q_{\curvearrowright}
    - 2 \sum_m q_m \log q_m \\
    &\quad - \sum_j \pi_j \log \pi_j
    + \sum_m (p_m + q_m) \log (p_m + q_m),
  \end{split}
  \label{eq:mapeq}
\end{equation}
where $q_m = \sum_{j \in m} \sum_{i \notin m} J_{ij}$ is module $m$'s exit
flow and $q_{\curvearrowright} = \sum_m q_m$. With $\pi$ held fixed, the
weights enter $L$ only through the exit flows, and
\begin{equation}
  M^{\ast}_m := \frac{\partial L}{\partial q_m}
    = \log \frac{q_{\curvearrowright} (p_m + q_m)}{q_m^2} \ge 0.
  \label{eq:Mstar}
\end{equation}

\paragraph{Intuition.} Leakage out of a module always costs bits.
$M^{\ast}_m$ is the marginal number of bits per unit of exit flow from $m$.
It is computed from $K$ module-level statistics and broadcast to module $m$:
this is the rule's neuromodulator.

\paragraph{The inverse problem.} Map-STDP pins the partition and minimizes
over the weights, averaged over the frames the environment produces:
\begin{equation}
  \min_W D(W), \qquad D(W) = \mathbb{E}_f \left[ L(J_f) \right].
  \label{eq:D}
\end{equation}

\paragraph{Check.} Differentiate~\eqref{eq:mapeq} term by term. The
derivative of $x \log x$ is $\log x + c$ for a constant $c$, and the
constants cancel ($c - 2c + c = 0$). The result is
$\log q_{\curvearrowright} - 2 \log q_m + \log(p_m + q_m)$. It is
non-negative because $q_{\curvearrowright} \ge q_m$ and
$p_m + q_m \ge q_m$.

\subsection{The structural rule}

Let $c_{ij}$ be the \emph{signed counter}: the change in $i$'s spikes
within the window caused by $j$'s kicks alone, with
$\mathbb{E}[c_{ij}] = \bar{n}_j D_{ij}$. The structural update is
\begin{equation}
  \begin{gathered}
  \Delta W_{ij} = -\lambda \left( g_{ij} - \bar{g}^{obs}_j \right) c_{ij}, \\
  g_{ij} = \frac{\partial D}{\partial J_{ij}},
  \qquad \bar{g}^{obs}_j = \sum_k T^{obs}_{kj} g_{kj}.
  \end{gathered}
  \label{eq:rule}
\end{equation}
For the map equation, $q_m$ is a sum of the $J_{ij}$ with $j \in m$ and
$i \notin m$, so
\begin{equation}
  g_{ij} = M^{\ast}_{m(j)} \mathbb{I}[i \notin m(j)], \qquad
  \bar{g}^{obs}_j = M^{\ast}_{m(j)} \bar{e}^{obs}_j,
\end{equation}
and the rule becomes
\begin{equation}
  \Delta W_{ij} = -\lambda M^{\ast}_{m(j)}
    \left( \mathbb{I}[i \notin m(j)] - \bar{e}^{obs}_j \right) c_{ij}.
  \label{eq:maprule}
\end{equation}

\paragraph{Two statistics, two jobs.} The baseline $\bar{e}^{obs}_j$
describes the walk, so it comes from pair counts, chance included. The step
must say how the walk changes when $W_{ij}$ changes, which is the direct
effect, so it comes from the signed counter. Using pair counts for the step
fails: they are mostly chance, and renormalization cancels them.

\paragraph{Sign pattern.} The factor
$\mathbb{I}[i \notin m(j)] - \bar{e}^{obs}_j$ is positive across modules and
negative within. The counter carries $\mathrm{sgn}_j$. Hence:

\begin{center}
\begin{tabular}{lcc}
  Synapse & Across modules & Within a module \\ \hline
  Excitatory & weakened & strengthened \\
  Inhibitory & strengthened & weakened \\
\end{tabular}
\end{center}

\subsection{Why the rule descends the description length}
\label{app:proof}

\paragraph{Local split.} While differentiating with respect to one weight,
hold fixed everything a single synapse does not directly touch: the flow
$\pi$, chance $a$, the indirect term $X$ and the gains. Then $W_{ij}$
changes only column $j$'s direct part,
$D_{kj} = \mathrm{sgn}_j h_k W_{kj} / d_j$ (through $W_{ij}$ itself and
through $d_j$).

\begin{lemma}[Lemma E]
\label{lem:E}
If $D$ depends on $W$ only through the link flows $J$, then under the local
split
\begin{equation}
  \begin{gathered}
  W_{ij} \frac{\partial D}{\partial W_{ij}}
  = \frac{\pi_j \mathrm{sgn}_j}{E_j}
    \left[ \tilde{\Lambda}_{ij} \left( g_{ij} - \bar{g}^{obs}_j \right)
           - T_{ij} C_j \right], \\
  C_j = \sum_k \tilde{\Lambda}_{kj} \left( g_{kj} - \bar{g}^{obs}_j \right).
  \end{gathered}
  \label{eq:lemmaE}
\end{equation}
\end{lemma}

\begin{proof}[Proof sketch]
Under the split,
$\partial E_{kj} / \partial W_{ij}
 = \mathrm{sgn}_j (h_k / d_j)(\delta_{ki} - T_{kj})$, and
$J_{kj} = \pi_j E_{kj} / E_j$. By the quotient rule,
\[
  \frac{\partial J_{kj}}{\partial W_{ij}}
  = \frac{\pi_j}{E_j} \left( \frac{\partial E_{kj}}{\partial W_{ij}}
    - T^{obs}_{kj} \frac{\partial E_j}{\partial W_{ij}} \right).
\]
Summing against $g_{kj}$, and using
$\partial E_j / \partial W_{ij} = \sum_k \partial E_{kj} / \partial W_{ij}$,
\begin{align*}
  \frac{\partial D}{\partial W_{ij}}
  &= \frac{\pi_j}{E_j} \sum_k \left( g_{kj} - \bar{g}^{obs}_j \right)
    \frac{\partial E_{kj}}{\partial W_{ij}} \\
  &= \frac{\pi_j \mathrm{sgn}_j}{E_j d_j}
    \Big[ h_i \left( g_{ij} - \bar{g}^{obs}_j \right) \\
  &\qquad - \sum_k h_k T_{kj} \left( g_{kj} - \bar{g}^{obs}_j \right) \Big].
\end{align*}
Multiplying by $W_{ij}$, with $W_{ij}/d_j = T_{ij}$ and
$h_i T_{ij} = \tilde{\Lambda}_{ij}$, gives~\eqref{eq:lemmaE}.
\end{proof}

\paragraph{Intuition.} $C_j$ is the same for every target $i$ of $j$, so
$T_{ij} C_j$ rescales column $j$ uniformly. That cannot change $T$, so it
cannot change $D$; hence $\sum_i W_{ij} \partial D / \partial W_{ij} = 0$.
The other factors read as follows:
\begin{itemize}
  \item $\pi_j / E_j$ is a per-neuron step size;
  \item $\mathrm{sgn}_j$ flips the direction for inhibitory synapses;
  \item $\tilde{\Lambda}_{ij}$ scales the step by how much $j$ actually moves $i$;
  \item $g_{ij} - \bar{g}^{obs}_j$ compares this transition's marginal cost
    with the average over $j$'s observed transitions.
\end{itemize}

\begin{theorem}[Theorem E]
\label{thm:E}
If the counter is unbiased,
$\mathbb{E}[c_{ij}] = T_f r_j \mathrm{sgn}_j \tilde{\Lambda}_{ij}$ over a
frame of duration $T_f$, then
\begin{equation}
  \begin{gathered}
  \mathbb{E}[\Delta W_{ij}]
  = -\lambda \zeta E_j W_{ij} \frac{\partial D}{\partial W_{ij}}
    - \lambda T_f r_j \mathrm{sgn}_j C_j T_{ij}, \\
  \Delta D = -\lambda \zeta \sum_{ij} E_j W_{ij}
    \left( \frac{\partial D}{\partial W_{ij}} \right)^2 \le 0,
  \end{gathered}
  \label{eq:thmE}
\end{equation}
with $\zeta = T_f \sum_k r_k$.
\end{theorem}

\begin{proof}[Proof sketch]
Solve~\eqref{eq:lemmaE} for
$\mathrm{sgn}_j \tilde{\Lambda}_{ij} (g_{ij} - \bar{g}^{obs}_j)$, substitute
it into $-\lambda (g_{ij} - \bar{g}^{obs}_j) \mathbb{E}[c_{ij}]$, and use
$\pi_j = r_j / \sum_k r_k$. To first order,
$\Delta D = \sum_{ij} (\partial D / \partial W_{ij}) \Delta W_{ij}$. The
column-rescaling term contributes
$\propto \sum_i W_{ij} \partial D / \partial W_{ij} = 0$.
\end{proof}

\paragraph{Intuition.} On average the rule is gradient descent on $D$,
preconditioned by the positive diagonal factor $E_j W_{ij}$ (which never has
to be computed), plus a column rescaling that cannot change $D$.

\paragraph{Flow-engine case.} With $a = X = 0$, excitation only and equal
gains, $T^{obs} = T$ and $C_j = 0$, and Lemma E reduces to
\begin{equation}
  \frac{\partial D}{\partial W_{ij}}
    = \frac{\pi_j}{d_j} \left( g_{ij} - \bar{g}_j \right),
  \qquad \bar{g}_j = \sum_k T_{kj} g_{kj}.
\end{equation}
The flow engine replaces the counter by its expectation $J_{ij}$, giving
$\mathbb{E}[\Delta W_{ij}] = -\lambda W_{ij} \partial D / \partial W_{ij}$:
gradient descent with each coordinate scaled by its own weight.

\subsection{Routing exemption}
\label{app:routing}

The policy is carried by cross-module flow: controller $\to$ action and
controller $\to$ latent. Applied to those synapses, the map term would drive
the controller's exit flow to zero and starve the action modules. The
routing pairs are therefore exempt: $g_{ba} = 0$ for $a = \mathcal{C}$ and
$b \in \mathcal{A} \cup \mathcal{L}$. The controller then acts as an input
layer, and the map term seals the latent and action modules.

\subsection{Task term}

The task is learned by a TD-modulated, action-gated eligibility. Take the
surrogate policy
$\tilde{P}(a) = J^{in}_a / \sum_{a' \in \mathcal{A}} J^{in}_{a'}$, with
$J^{in}_b = \sum_{i \in b} \sum_j J_{ij}$. Its score with respect to
$J_{ij}$ depends only on the module of $i$:
\begin{equation}
  h_b = \frac{\mathbb{I}[b = a^{\ast}]}{J^{in}_{a^{\ast}}}
      - \frac{\mathbb{I}[b \in \mathcal{A}]}{\sum_{a' \in \mathcal{A}} J^{in}_{a'}},
  \qquad \bar{h}_j = \sum_k T_{kj} h_{m(k)},
\end{equation}
where $a^{\ast}$ is the chosen action. The eligibility trace and the update
are
\begin{equation}
  e_{ij} \leftarrow \left( 1 - \tfrac{1}{\tau_e} \right) e_{ij}
    + c_{ij} \left( h_{m(i)} - \bar{h}_j \right),
  \qquad
  \Delta W_{ij} = \eta \delta_f e_{ij},
  \label{eq:task}
\end{equation}
with the TD error and a linear critic trained by normalized LMS:
\begin{equation}
  \begin{gathered}
  \delta_f = r_f + \gamma V_{f+1} - V_f, \qquad V_f = u \cdot x_f, \\
  u \leftarrow u + \eta_V \delta_f \frac{x_f}{\lVert x_f \rVert^2}.
  \end{gathered}
  \label{eq:td}
\end{equation}
In the flow engine, by the same Lemma, the increment of~\eqref{eq:task} is
$W_{ij} \partial \log \tilde{P}(a^{\ast}) / \partial W_{ij}$: a one-hop
policy-gradient score.

\subsection{Homeostasis}

A presynaptic homeostasis term holds each neuron's output weight near 1:
\begin{equation}
  \Delta W_{ij} = -\epsilon_h (d_j - 1) \frac{\pi_j W_{ij}}{d_j}.
  \label{eq:homeo}
\end{equation}
It rescales column $j$ uniformly, so it leaves $T$, the walk and $D$
unchanged. The full per-frame update, Eq.~\eqref{eq:full}, is the sum of the
task, structure and homeostasis terms.

\subsection{Computing the counter in each engine}
\label{app:counter}

Each engine uses the same rule. What differs is how $c_{ij}$ is computed.

\paragraph{Branching.} The neuron is
$u_i(t+1) = \sum_j b_{ij} x_j(t) + \text{noise}$, which spikes with
probability $p(u) = \mathrm{clip}(u / 2\theta, 0, 1)$. The kick
$b_{ij} = \mathrm{sgn}_j w_0 T_{ij}$ is output-normalized. Removing $j$'s
kick changes $i$'s firing probability by a known amount, so the counter is
exact:
\begin{equation}
  c_{ij} = \sum_t x_j(t)
    \left[ p\left( u_i(t+1) \right) - p\left( u_i(t+1) - b_{ij} \right) \right].
\end{equation}

\paragraph{LIF and QIF.} A kick of size $b$ is an input pulse of area
$\tau b$. In linear response it changes $i$'s spike count by
$\hat{G}_i \tau b$, where $\hat{G}_i$ is $i$'s gain (the slope of its rate
transfer). So
\begin{equation}
  c_{ij} = \tau \hat{G}_i b_{ij} n_j,
\end{equation}
an outer product of the presynaptic spike count and the postsynaptic gain.
For LIF the gain comes from the Siegert rate formula, or from the exact
Markov transfer of the discretized neuron. For QIF it comes from regressing
the rate on the injected noise.

\subsection{Stability}
\label{app:stability}

The walk always exists, but the counters rely on linear response around a
stable operating point. With $\Lambda$ the signed offspring matrix (extra
spikes of $i$ per spike of $j$), the conditions are
\begin{equation}
  \begin{gathered}
  \text{branching (map):} \ \max_k \lvert \lambda_k(\Lambda) \rvert < 1, \\
  \text{LIF, QIF (continuous):} \ \max_k \mathrm{Re} \lambda_k(\Lambda) < 1.
  \end{gathered}
\end{equation}
The network monitors them through a module-level offspring table estimated
from the counters.

\subsection{Map-STDP communities in the TD-STDP backbone}
\label{app:backbone}

In the hybrid model, the actor of the TD-STDP actor-critic of Chung and
Kozma~\cite{chung2020} is organized into modules:
\begin{itemize}
  \item \textbf{Controller.} The controller $\mathcal{C}$ is the Fourier
    input population.
  \item \textbf{Action communities.} Each action's excitatory actor population
    is one module.
  \item \textbf{Recurrence.} Recurrent synapses $R$ connect all actor
    excitatory neurons, with output-normalized kicks
    $b_{ij} = \chi_{rec} R_{ij} / d_j$.
  \item \textbf{Routing.} The input $\to$ actor weights are exempt routing,
    learned by the TD-STDP rule alone.
  \item \textbf{Where the map term acts.} The structural term acts on $R$
    only, with the LIF counter $c_{ij} = \tau \hat{G}_i b_{ij} n_j$
    accumulated per window.
\end{itemize}
The action is sampled from a softmax over community rates,
$p_k \propto \exp(\alpha_a \bar{\rho}_k)$. This is equivalent to a race of
escape-noise decision units with constant hazards
$\exp(\alpha_a \bar{\rho}_k)$, in which the first unit to spike
wins~\cite{yellott1977}. Each unit's local score
$\mathbb{I}[k = a] - \Lambda_k(t^{\ast})$, with $\Lambda_k$ its integrated
hazard up to the decision time $t^{\ast}$, has expectation
$A_k - p_k$ given the winner, where $A$ is the one-hot chosen action.
So the community credit $h_b \propto A_b - p_b$ of the backbone is available
locally. The task term on $R$ uses the one-hop credit
$h_{m(i)} - \bar{h}_j$ of~\eqref{eq:task}.

\section{Random Walk Entropy and SNN Energy Usage}

In this section we discuss in more depth how entropy of a random walk abstraction of SNN inference relates to the energy usage of the network (Fig.~\ref{fig:energy-px} and Table~\ref{tab:energy-walk-corr}).

\subsection{Energy Usage}
In this supplementary study, we measure energy usage through spiking operations (SOPs). An SOP is one spike delivered across one synapse. Each spike is weighted by the number of outgoing synapses of the neuron that fired it: 

$$\text{SOPs} = \sum_j n_j \cdot |\lbrace i : W_{ij} > 2 w_{\min} \rbrace|$$

Here $n_j$ is the spike count of neuron $j$. The count only includes synapses above twice the weight floor ($w_{\min} = 10^{-3}$), the “effective” synapses. Synapses the rule has pushed down to the floor are treated as pruned, so an SOP is only counted where a synapse effectively exists. SOPs are computed on frozen weight models without any continual learning.

\subsection{$L_W$ and $L_{obs}$}
There are two measurements of info-theoretic entropy in the random-walk abstraction: the map equation of the observed walk ($L_{\mathrm{obs}}$) and the map equation of the walk on the trained weights ($L_W$). 
Both description lengths are the map equation of a walk with visit rates $\pi$ and transitions $T$ over the modules $m$ (see Equation~\eqref{eq:map-equation}).
They differ in the walk they use:
\begin{align}
L_W &= L(\pi^W, T^W), & T^W_{ij} &= W_{ij} \Big/ \textstyle\sum_k W_{kj}, \\
& & \pi^W &= (1 - \alpha) T^W \pi^W + \tfrac{\alpha}{N} \mathbf{1}, \quad \alpha = 0.2, \notag \\
L_{\mathrm{obs}} &= \bigl\langle L(\pi_f, T^{\mathrm{obs}}_f) \bigr\rangle_f, & T^{\mathrm{obs}}_{ij} &=
  \kappa_{ij} \Big/ \textstyle\sum_k \kappa_{kj}, \\
& & \pi_{f,j} &= n_j \Big/ \textstyle\sum_k n_k. \notag
\end{align}
$L_W$ is computed from the weights alone (the recurrent weights $R$ for TD-STDP) under uniform teleportation. $L_{\mathrm{obs}}$
is computed from spikes and averaged over frames $f$. In it, $\kappa_{ij}$ counts spikes of $i$ within the engine's window after
a spike of $j$, and $n_j$ is the spike count of neuron $j$ in the frame.

\subsection{Info-theoretic entropy and downstream energy use}
No experiment was run on the flow model as there are no spiking operations in the flow model. The full correlation table can be found in Table~\ref{tab:energy-walk-corr}. Overall, there is general agreement between energy usage and description length (Spearman correlations: +0.64 overall SOPs vs $L_{obs}$, +0.53 overall SOPs vs $L_W$), showing strong evidence for lower description length yielding lower energy usage. Moreover, there is even a weak negative correlation between performance and the two description length quantities (-0.19 Return vs $L_{obs}$ and vs $L_W$), providing a small amount of evidence for Hinton's compressibility argument. However, the agreement between info-theoretic entropy via description length and downstream energy usage, as well as the two measures of description length, depends on the model and the task.

The correlation between SOPs and $L_{obs}$ becomes weak and even goes slightly negative in the full modularity regime in the TD-STDP backbone on Acrobot (-0.13 Spearman correlation). The correlation between SOPs and $L_W$, and between $L_{obs}$ and $L_W$ drops heavily on LunarLander, indicating that preemptive info-theoretic analysis holds weaker on this task. 

\begin{figure}[!ht]
\centering
\includegraphics[width=\textwidth]{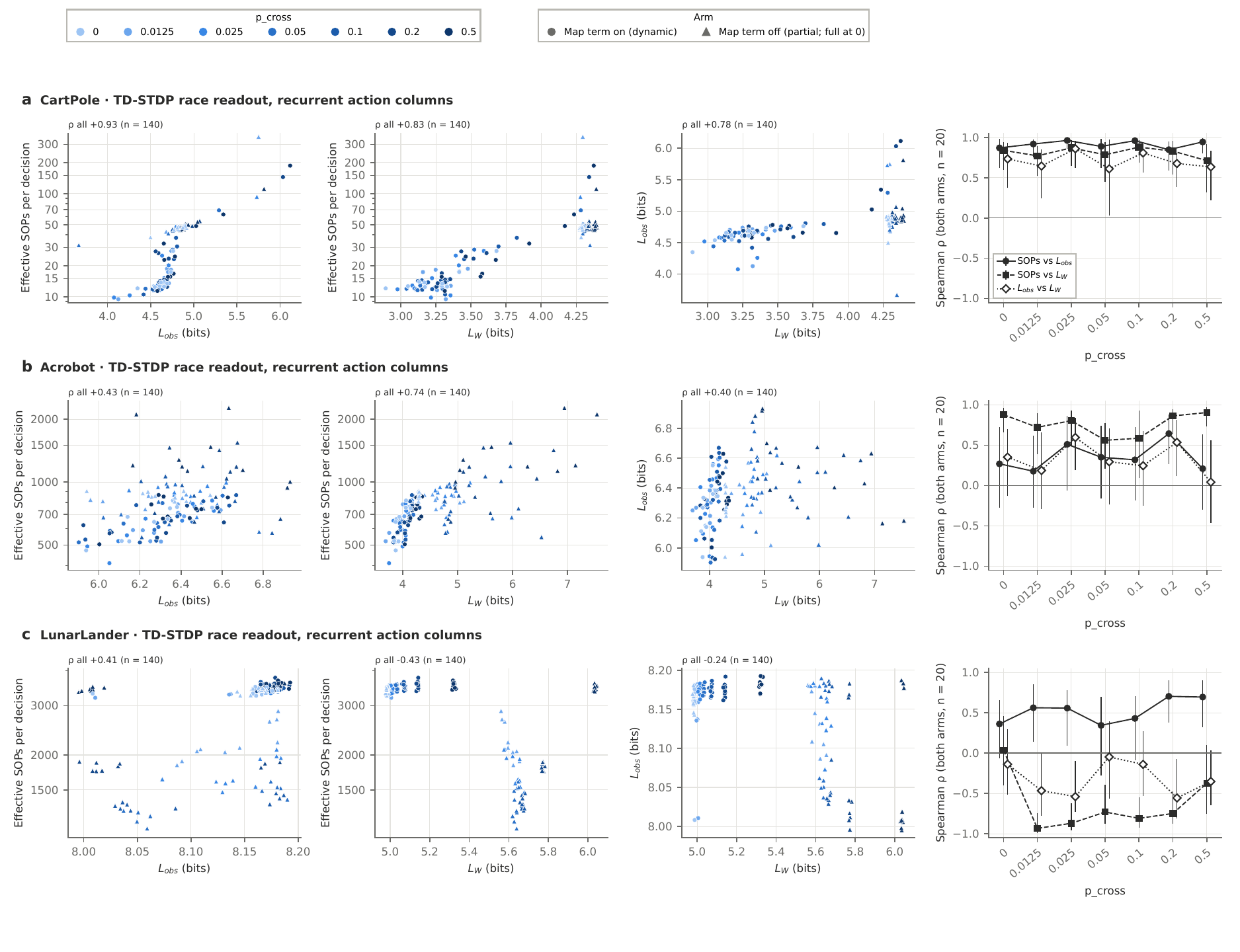}
\caption{Energy use and description length across the inter-module connectivity sweep in the TD-STDP actor-critic backbone, for (a)~CartPole, (b)~Acrobot and (c)~LunarLander. Each point is one final network (7 values of $p_{cross}$ $\times$ 2 map-term settings $\times$ 20 seeds, $n = 140$); color gives $p_{cross}$ and shape the map-term setting (dynamic with the term on, partial with it off, full at $p_{cross} = 0$ with it off). Columns 1--3: effective synaptic operations (SOPs) per decision against $L_{obs}$ and $L_W$, and $L_{obs}$ against $L_W$, with the Spearman $\rho$ over all points above each panel. Column 4: Spearman $\rho$ within each $p_{cross}$ value ($n = 20$ per setting), with bootstrap intervals.}
\label{fig:energy-px}
\end{figure}

In Figure~\ref{fig:energy-px}, showing the correlation between SOPs and the two description lengths across $p_{cross}$ values, we see a similar trend. There is a positive correlation between SOPs and $L_{obs}$ across all three tasks (CartPole: +0.93, Acrobot: +0.43, LunarLander: +0.41). The correlation between all three quantities are strongest in CartPole (SOPs vs $L_{obs}$: +0.93, SOPs vs $L_W$: +0.83, $L_{obs}$ vs $L_W$: +0.78). And LunarLander holds the weakest correlation, with theoretical description length analysis even serving counterintuitively (SOPs vs $L_{obs}$: +0.41, SOPs vs $L_W$: -0.43, $L_{obs}$ vs $L_W$: -0.24). Across the $p_{cross}$ values, in the plots on the rightmost column, correlations hold positive consistently in CartPole and Acrobot, and while SOPs and $L_{obs}$ hold positive, the correlations with $L_W$ drop negative consistently. This opens the question of what attributes the LunarLander task has for preemptive info-theoretic analysis to hold so counterintuitively. This question is explored in the following subsection.

\begin{table}[t]
\centering
\small
\caption{Spearman correlation between energy use, random-walk description length and performance of the final networks: effective synaptic operations (SOPs) per decision, the map equation of the observed walk ($L_{\mathrm{obs}}$) and of the walk on the trained weights ($L_W$), and the mean return of the frozen final network. $\rho$ with its bootstrap 95\% interval (2000 resamples over seeds) below; $^{*}$ $p < 0.05$. Map-STDP v4: per spiking engine, pooled over the three modularity regimes. TD-STDP: all regimes pooled, then within each regime. All settings: each variable ranked within its setting (three v4 engines, three TD-STDP tasks) and pooled; bootstrap stratified by setting. ---: every frozen TD-STDP CartPole network reaches the 500 ceiling, so its return is constant; that setting is left out of the all-settings return columns ($n = 300$).}
\label{tab:energy-walk-corr}
\begin{tabular}{lrccccc}
\toprule
 & & \multicolumn{2}{c}{SOPs vs} & $L_{\mathrm{obs}}$ vs & \multicolumn{2}{c}{Return vs} \\
\cmidrule(lr){3-4} \cmidrule(lr){5-5} \cmidrule(lr){6-7}
Group & $n$ & $L_{\mathrm{obs}}$ & $L_W$ & $L_W$ & $L_{\mathrm{obs}}$ & $L_W$ \\
\midrule
\multicolumn{7}{l}{\textit{Map-STDP v4, CartPole}} \\
Branching & 60 & $+0.61^{*}$ & $+0.71^{*}$ & $+0.49^{*}$ & $-0.49^{*}$ & $-0.19$ \\
 & & {\scriptsize $[+0.35, +0.77]$} & {\scriptsize $[+0.52, +0.81]$} & {\scriptsize $[+0.21, +0.67]$} & {\scriptsize $[-0.70, -0.24]$} & {\scriptsize $[-0.44, +0.08]$} \\[2pt]
QIF & 60 & $+0.95^{*}$ & $+0.93^{*}$ & $+0.92^{*}$ & $-0.09$ & $-0.11$ \\
 & & {\scriptsize $[+0.90, +0.97]$} & {\scriptsize $[+0.87, +0.96]$} & {\scriptsize $[+0.86, +0.95]$} & {\scriptsize $[-0.35, +0.18]$} & {\scriptsize $[-0.37, +0.17]$} \\[2pt]
LIF & 60 & $+0.91^{*}$ & $+0.91^{*}$ & $+0.92^{*}$ & $-0.50^{*}$ & $-0.45^{*}$ \\
 & & {\scriptsize $[+0.86, +0.94]$} & {\scriptsize $[+0.86, +0.93]$} & {\scriptsize $[+0.86, +0.95]$} & {\scriptsize $[-0.68, -0.28]$} & {\scriptsize $[-0.64, -0.23]$} \\[2pt]
\midrule
\multicolumn{7}{l}{\textit{TD-STDP, CartPole}} \\
All regimes & 60 & $+0.88^{*}$ & $+0.72^{*}$ & $+0.71^{*}$ & --- & --- \\
 & & {\scriptsize $[+0.78, +0.94]$} & {\scriptsize $[+0.53, +0.84]$} & {\scriptsize $[+0.52, +0.84]$} &  &  \\[2pt]
Dynamic & 20 & $+0.83^{*}$ & $+0.55^{*}$ & $+0.66^{*}$ & --- & --- \\
 & & {\scriptsize $[+0.56, +0.95]$} & {\scriptsize $[+0.11, +0.85]$} & {\scriptsize $[+0.20, +0.93]$} &  &  \\[2pt]
Partial & 20 & $+0.65^{*}$ & $-0.23$ & $-0.29$ & --- & --- \\
 & & {\scriptsize $[+0.24, +0.87]$} & {\scriptsize $[-0.61, +0.21]$} & {\scriptsize $[-0.73, +0.24]$} &  &  \\[2pt]
Full & 20 & $+0.65^{*}$ & $+0.25$ & $+0.02$ & --- & --- \\
 & & {\scriptsize $[+0.24, +0.88]$} & {\scriptsize $[-0.22, +0.65]$} & {\scriptsize $[-0.52, +0.55]$} &  &  \\[2pt]
\midrule
\multicolumn{7}{l}{\textit{TD-STDP, Acrobot}} \\
All regimes & 60 & $+0.02$ & $+0.38^{*}$ & $+0.07$ & $-0.04$ & $-0.10$ \\
 & & {\scriptsize $[-0.24, +0.28]$} & {\scriptsize $[+0.11, +0.63]$} & {\scriptsize $[-0.19, +0.31]$} & {\scriptsize $[-0.33, +0.24]$} & {\scriptsize $[-0.34, +0.14]$} \\[2pt]
Dynamic & 20 & $+0.32$ & $+0.46^{*}$ & $+0.61^{*}$ & $+0.02$ & $+0.26$ \\
 & & {\scriptsize $[-0.21, +0.77]$} & {\scriptsize $[+0.03, +0.80]$} & {\scriptsize $[+0.24, +0.82]$} & {\scriptsize $[-0.44, +0.45]$} & {\scriptsize $[-0.24, +0.66]$} \\[2pt]
Partial & 20 & $+0.18$ & $-0.15$ & $-0.64^{*}$ & $+0.13$ & $-0.30$ \\
 & & {\scriptsize $[-0.32, +0.61]$} & {\scriptsize $[-0.59, +0.41]$} & {\scriptsize $[-0.87, -0.21]$} & {\scriptsize $[-0.36, +0.59]$} & {\scriptsize $[-0.71, +0.20]$} \\[2pt]
Full & 20 & $-0.13$ & $+0.18$ & $+0.06$ & $-0.39$ & $-0.02$ \\
 & & {\scriptsize $[-0.59, +0.40]$} & {\scriptsize $[-0.34, +0.60]$} & {\scriptsize $[-0.41, +0.51]$} & {\scriptsize $[-0.75, +0.08]$} & {\scriptsize $[-0.52, +0.41]$} \\[2pt]
\midrule
\multicolumn{7}{l}{\textit{TD-STDP, LunarLander}} \\
All regimes & 60 & $+0.49^{*}$ & $-0.47^{*}$ & $-0.01$ & $+0.19$ & $-0.12$ \\
 & & {\scriptsize $[+0.25, +0.69]$} & {\scriptsize $[-0.69, -0.17]$} & {\scriptsize $[-0.30, +0.30]$} & {\scriptsize $[-0.09, +0.43]$} & {\scriptsize $[-0.37, +0.16]$} \\[2pt]
Dynamic & 20 & $+0.11$ & $+0.29$ & $+0.26$ & $+0.07$ & $-0.09$ \\
 & & {\scriptsize $[-0.44, +0.55]$} & {\scriptsize $[-0.14, +0.65]$} & {\scriptsize $[-0.27, +0.71]$} & {\scriptsize $[-0.50, +0.54]$} & {\scriptsize $[-0.52, +0.41]$} \\[2pt]
Partial & 20 & $+0.58^{*}$ & $-0.69^{*}$ & $-0.39$ & $+0.65^{*}$ & $-0.79^{*}$ \\
 & & {\scriptsize $[+0.13, +0.79]$} & {\scriptsize $[-0.87, -0.35]$} & {\scriptsize $[-0.72, +0.12]$} & {\scriptsize $[+0.22, +0.88]$} & {\scriptsize $[-0.93, -0.50]$} \\[2pt]
Full & 20 & $+0.26$ & $+0.49^{*}$ & $-0.14$ & $+0.18$ & $-0.08$ \\
 & & {\scriptsize $[-0.24, +0.69]$} & {\scriptsize $[+0.03, +0.79]$} & {\scriptsize $[-0.60, +0.37]$} & {\scriptsize $[-0.30, +0.60]$} & {\scriptsize $[-0.54, +0.41]$} \\[2pt]
\midrule
\textbf{All settings} & 360 & $+0.64^{*}$ & $+0.53^{*}$ & $+0.52^{*}$ & $-0.19^{*}$ & $-0.19^{*}$ \\
 & & {\scriptsize $[+0.56, +0.71]$} & {\scriptsize $[+0.45, +0.60]$} & {\scriptsize $[+0.43, +0.59]$} & {\scriptsize $[-0.30, -0.07]$} & {\scriptsize $[-0.30, -0.08]$} \\
\bottomrule
\end{tabular}
\end{table}

\subsection{The Special Case of LunarLander}

Synaptic operations (SOPs) here track the number of surviving synapses, not activity. Within each task, column spikes per decision barely change across regimes: about 5 in CartPole, 92 in Acrobot and 211--223 in LunarLander, since the rate-matching settings of these studies hold firing rates fixed. SOPs therefore vary only through the effective fan-out, the number of a neuron's recurrent synapses still above threshold (Table~\ref{tab:ll-fanout}).

\begin{table}[h]
\centering
\small
\caption{Effective fan-out (recurrent synapses per neuron above $2 w_{\min}$) of the final networks, mean over seeds.}
\label{tab:ll-fanout}
\begin{tabular}{lccc}
\toprule
Task & Dynamic & Full & Partial \\
\midrule
CartPole & 3.3 & 9.5 & 9.5 \\
Acrobot & 7.1 & 9.6 & 8.3 \\
LunarLander & 15.6 & 15.6 & \textbf{6.2} \\
\bottomrule
\end{tabular}
\end{table}

$L_W$, in contrast, measures how much of the weight crosses between columns. The two usually move together because one process drives both: the map term removes cross-column synapses, lowering the synapse count and $L_W$ at once. Without the map term, each task treats the cross-column synapses differently (Table~\ref{tab:ll-drift}).

\begin{table}[h]
\centering
\small
\caption{Recurrent weights of the partial arm (map term off) from the start to the end of training, mean over seeds. Within-column synapses: per neuron, above $2 w_{\min}$. Within-column share: fraction of the total recurrent weight inside columns.}
\label{tab:ll-drift}
\begin{tabular}{lccc}
\toprule
Task & Mean cross-column weight & Within-column synapses & Within-column share \\
\midrule
CartPole & $0.094 \rightarrow 0.001$ & $9.5 \rightarrow 9.5$ & $0.91 \rightarrow 1.00$ \\
Acrobot & $0.084 \rightarrow 0.187$ & $9.5 \rightarrow 6.9$ & $0.83 \rightarrow 0.62$ \\
LunarLander & $0.048 \rightarrow 0.140$ & $\mathbf{15.6 \rightarrow 4.5}$ & $0.77 \rightarrow \mathbf{0.31}$ \\
\bottomrule
\end{tabular}
\end{table}

In CartPole the task rule removes the cross-column synapses by itself within about 200 episodes, so the partial arm ends up equal to the full one. In Acrobot the cross-column weights double, but the columns survive. In LunarLander the cross-column weights triple and most within-column synapses fall below threshold, mainly between episodes 2{,}500 and 5{,}500. The final within-column share of 0.31 is below the 0.77 of the random initial wiring, so the network ends up less modular than chance.

In LunarLander's partial arm, then, the pruned synapses are the within-column ones, and fewer synapses go with more cross-column flow. This makes SOPs and $L_W$ negatively correlated ($\rho = -0.47$ over all regimes, $-0.69$ within the partial arm). With the map term on, the cross-column synapses go and all 15.6 within-column synapses stay, which is why the dynamic and full arms end up with the same SOPs and $L_W$.

Why the within-column synapses decay is not yet established. Candidates are the rate cost, homeostasis, or the task rule over LunarLander's much longer training (10{,}000 episodes, against 400 for CartPole and 500 for Acrobot).

A fixed partition thus appears too rigid for this task, and unconstrained coupling too loose. The map term alone has limitations expressing a middle ground: its modulator $g_{ba} = M^{\ast}_a \mathbb{I}[b \neq a]$ charges every exit from column $a$ the same marginal cost, whatever the target column $b$. A natural extension is an adaptive $K \times K$ module-pair table learned for instance from the task's credit, so that useful directed links between action columns are kept while the rest are pruned. We leave this to future work.

\end{document}